# Metaprobability and Dempster-Shafer
## In Evidential Reasoning

Robert M. Fung
Chee-Yee Chong

Advanced Information & Decision Systems
201 San Antonio Circle, Suite 286
Mountain View, California 94040

## 1. Introduction

Evidential reasoning in expert systems has often used ad-hoc uncertainty calculi. Although it is generally accepted that probability theory provides a firm theoretical foundation, researchers have found some problems with its use as a workable uncertainty calculus. Among these problems are representation of ignorance, consistency of probabilistic judgements, and adjustment of a priori judgements with experience. The application of metaprobability theory to evidential reasoning is a new approach to solving these problems. We use the Dempster-Shafer theory, an alternative theory of evidential reasoning to judge metaprobability theory as a theory of evidential reasoning. This paper will compare how metaprobability theory and Dempster-Shafer theory handle the adjustment of judgements with experience.

Section 2 and 3 describe the basics of the metaprobability and Dempster-Shafer theories. Metaprobability theory deals with higher order probabilities applied to evidential reasoning. Dempster-Shafer theory is a generalization of probability theory which has evolved from a theory of upper and lower probabilities.

Section 4 describes a thought experiment and the metaprobability and Dempster-Shafer analysis of the experiment. The thought experiment focuses on forming beliefs about a die from evidence accrued from two sensors: an odd-even sensor, and a large-small sensor. For a large number of tosses, the odd-even sensor sees half the tosses come up even, and the other half come up odd. For a different set of tosses of the same number, the large-small sensor sees half the tosses come up large and the other half small. Based on these two pieces of evidence, what should be the beliefs about the die?

## 2. Metaprobability Theory

Metaprobability theory deals with probability measures on the space of first-order probability distributions. In turn, the first-order distributions are defined over some domain state-space. In the field of probability theory, metaprobability has been known as "higher order probability" or "hierarchical probability". However, it is more appropriate to refer to the application of higher order probability to evidential reasoning as metaprobability because the order of the probability distribution corresponds to the order of the meta-level at which the probability theory is attempting to provide a quantifiable model of the evidential accrual process.

Many theorists have considered it only briefly due to the lack of practical applications to motivate the use of metaprobabilities as well as a lack of computational resources to implement them. We believe that evidential reasoning is an appropriate application of metaprobabilities and that the computational problem can be overcome with current technology.

Figure 1 shows gives an example. The domain state-space is the set of (heads, tails). The space of consistent probability distributions consists of those such that $p(\text{heads}) + p(\text{tails}) = 1$. Two different metaprobability distributions are shown. The flat distribution represents extreme ignorance about this situation. The other distribution represents knowledge that the coin is most probably "fair".

There are no theoretical questions at the level of the mathematical formalism since metaprobability theory has the same mathematical foundations as probability theory. "Operations with probabilities of the second level are determined entirely by the rules of operations with probabilities of the first level" [Reichenbach 1949]. However, there are important philosophical issues and methodological issues. The philosophical issues are concerned mainly with the psychological and logical validity of metaprobabilities. The methodological issues are concerned mainly with controlling the combinatorics of metaprobability state spaces.

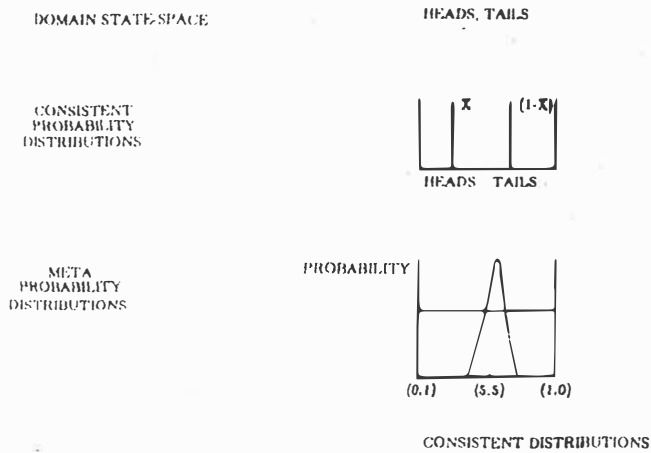

Figure 1: Metaprobability Examples

Metaprobabilistic updating of beliefs given evidence is based on a straight-forward application of Bayes rule:

$$MLPD\{p \mid E, S\} = K \cdot MLPD\{E \mid p, S\} \cdot MLPD\{p \mid S\}$$

p - first order probability distribution
E - experimental evidence
S - prior information

$$K = \frac{1}{MLPD\{E \mid S\}}$$

$MLPD\{p \mid S\}$ denotes the prior meta-level distribution and $MLPD\{p \mid E, S\}$ denotes the posterior distribution. The $MLPD\{E \mid p, S\}$ is a measure of the likelihood of a particular body of evidence, given a particular underlying first-order distribution. The evaluation of this term is simple. For example, if $E$ is an observed odd throw, the sample space is the sides of a die, and we are given a particular probability density $p = \{p(1), p(2), p(3), p(4), p(5), p(6)\}$, then the value of $MLPD\{$odd throw $\mid p, S\}$ is just $p(1) + p(3) + p(5)$.

77

## 3. Dempster-Shafer Theory

Development of the Dempster-Shafer theory was started by Arthur Dempster [Dempster 1967] and extended by Glenn Shafer [Shafer, 1976]. Many artificial intelligence researchers have turned to this theory in the hope of avoiding the pitfalls of using probability theory in evidential reasoning. Dempster-Shafer theory is a generalization of probability theory with its roots in a theory of upper and lower probabilities. Consequently, some of the important ideas in the theory can be thought of in terms of upper and lower probabilities. We give a brief description of the theory.

Let $\Theta$ be a set of mutually exclusive and exhaustive propositions about a domain. $\Theta$ is called the frame of discernment. Let $2^\Theta$ denote the set of all subsets of $\Theta$. Subsets of $\Theta$ are the propositions with which the theory is concerned. As a matter of convention, $A$ and $B$ will denote subsets of a frame of discernment.

We define three functions: the basic probability assignment $m$, the belief function $Bel$, and the plausibility function $Pl$. These functions have the same domain and range. Their domain is $2^\Theta$ and their range is $[0,1]$.

A "basic probability assignment" ($m$) must meet these constraints:

$$1) \quad m(\emptyset) = 0 \qquad 2) \quad \sum_{A \subset \Theta} m(A) = 1$$

The "belief" function can be defined in terms of a basic probability assignment $m$ by:

$$Bel(A) = \sum_{B \subset A} m(B)$$

The "plausibility" function can be defined in terms of a belief function $Bel$ by:

$$Pl(A) = 1 - Bel(\overline{A}) \quad \text{where } \overline{A} \text{ is the complement of } A.$$

Each of the three functions carries the same information. That is, there is a unique one-to-one transformation between any of the functions. However each function has a unique interpretation. The "basic probability assignment" of a subset of $\Theta$ can be interpreted as the probability mass constrained to stay in the subset. The "belief" of a subset of $\Theta$ can be interpreted as the measure of the lower probability of the subset, that is the minimum probability mass in the subset. And the "plausibility" of a subset of $\Theta$ is the upper probability of the subset, that is the maximum probability mass in the subset.

A focal element $A$ of a frame of discernment $\Theta$ is a subset of $\Theta$ such that $m(A) > 0$. The union of the focal elements of a belief function is called its core.

Frames of discernment can be made to distinguish finer and coarser concepts by processes called refining and coarsening. A frame of discernment $\Omega$ is a refinement of another frame $\Theta$ if there exists refining function $\omega$ such that:

$$1) \quad \omega(\{\theta\}) \neq \emptyset \text{ for } \forall \theta \in \Theta$$
$$2) \quad \omega(\{\theta\}) \cap \omega(\{\theta'\}) = \emptyset \text{ if } \theta \neq \theta'$$
$$3) \quad \bigcup_{\theta \cdot \Theta} \omega(\{\theta\}) = \Omega$$

Given a frame of discernment $\Omega$, a frame of discernment $\Theta$, a refining function $\omega$ from $\Theta$ to $\Omega$, and a belief function $Bel$ over $\Theta$ we define $Bel_0$ to be the vacuous extension of $Bel$ :

$$Bel(A) = \max_{B \subset \Theta, \omega(B) \subset A} Bel_0(B)$$

The vacuous extension of a belief function can be interpreted as the belief function in the refined frame which places only the constraints explicitly required by the belief function in the coarsened frame.

All belief functions are classified into two classes: support functions, and quasi-support functions. Quasi-support functions are those belief functions which have all of their probability masses on sets which are mutually exclusive. Quasi-support functions contain the set of probability mass functions. Support functions are those belief functions which Shafer believes "constitutes the subclass of belief functions appropriate for the representation of evidence". Support functions are defined by those belief functions whose core has a positive mass function. What this means is that every subset in the frame of discernment has a lower probability $Bel$ which is strictly less than its upper probability $Pl$.

Given two belief functions over the same frame of discernment but based on distinct bodies of evidence, Dempster's Rule of Combination can be used to compute a new belief function based on the combined evidence. Dempster's Rule of Combination is defined by:

$$m(A) = k \cdot \sum_{(B_1 \cap B_2 = A)} m_1(B_1) \cdot m_2(B_2)$$

## 4. Experiment

We present a thought experiment designed to compare how metaprobability theory and Dempster-Shafer theory update beliefs with evidence. The experiment is as follows:

"We are given a normal six-sided die and two sensors. One sensor can sense whether the top face of the die is odd or even. The other sensor can sense whether the top face of the die has four or more spots (large) or has three or less spots (small). We throw the die a large number of times with the odd-even sensor watching. The sensor reports that in exactly half the throws the die came up even and in the other half it came up odd. We throw the die the same number of times as before with the large-small sensor watching. The sensor reports that exactly half of the throws were large and half were small. What are our beliefs about the outcome of next throw of the die given we can distinguish between the six outcomes?"

Metaprobability Results

We have calculated the results of a simulated die throwing experiment using the metaprobability updating rule. We assumed a uniform prior $MLPD$, i.e. given any two first order probability distributions $p$ and $q$, we initially assume that $MLPD(p) = MLPD(q)$. We use a simple updating rule based on Bayes Rule to update the evidence. Figure 2 shows the updating process.

In order to simulate probability distributions in a machine environment, it is necessary to choose a discretization of the interval [0,1]. We initially chose a coarse quantization into sixths for the first order probability distributions. That is, every probability mass is restricted to the values 0, 1/6, 1/3, 1/2, 2/3, 5/6, or 1. Discretizations by twelfths and eighteens given the same problem, priors, and evidence, yielded analogous results to those described below.

The experimental results show that in the limit (i.e. the number of tosses approaches infinity) the updated $MLPD$ will consist of the distributions which meet the obvious "constraints" suggested by the evidence on odd-even and small-large probabilities. These constraints are:

$p(1) + p(2) + p(3) = .5 \qquad p(1) + p(3) + p(5) = .5$

$p(4) + p(5) + p(6) = .5 \qquad p(2) + p(4) + p(6) = .5$

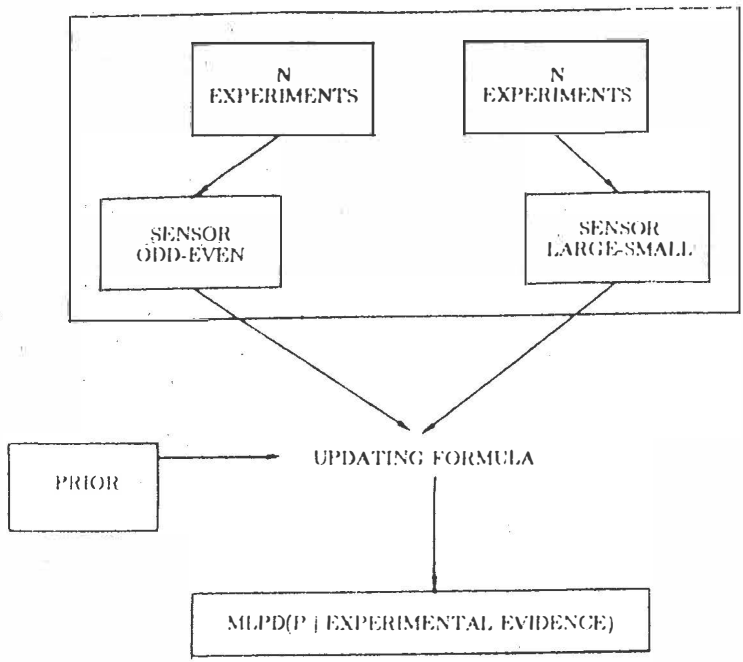

Figure 2: Metaprobability Analysis

With respect to the chosen discretization (i.e. 0, 1/6, ..., 5/6, 1), there are thirty first-order distributions for which this holds. A few examples are given below. (Note that $p(1) + p(3)$ is always equal to $p(4) + p(6)$.) A given row in the list is to be interpreted as the values for a first-order distribution.

sample space

| 1 | 2 | 3 | 4 | 5 | 6 |
|---|---|---|---|---|---|
| 3/6 | 0 | 0 | 3/6 | 0 | 0 |
| 2/6 | 0 | 1/6 | 2/6 | 0 | 1/6 |
| 1/6 | 2/6 | 0 | 1/6 | 2/6 | 0 |
| 0 | 3/6 | 0 | 0 | 3/6 | 0 |

Of the thirty distributions, sixteen distributions meet the constraint $p(1) + p(3) = p(4) + p(6) = .5$. Nine of the distributions meet the constraint $p(1) + p(3) = 2/6$, four meet the constraint $p(1) + p(3) = 1/6$, and one distribution meets the constraint $p(1) + p(3) = 0$.

## Dempster-Shafer Results

We have determined the results of the experiment using the Dempster-Shafer approach. The approach we take is first to use the statistical estimation technique presented by Shafer [Shafer, 1975] on the frames of discernment {odd, even} and {small, large} to create belief functions for the two experiments. The two belief functions are then refined to the common frame of discernment {1 2 3 4 5 6} and combined using Dempster's Rule of Combination to produce a belief function which is based on the combined evidence. Figure 3 shows the process described.

The first step is the statistical estimation technique on the frames of discernment {odd, even} and {small, large}. In the limit, (i.e. as the number of trials goes to infinity) the basic probability assignments converge to:

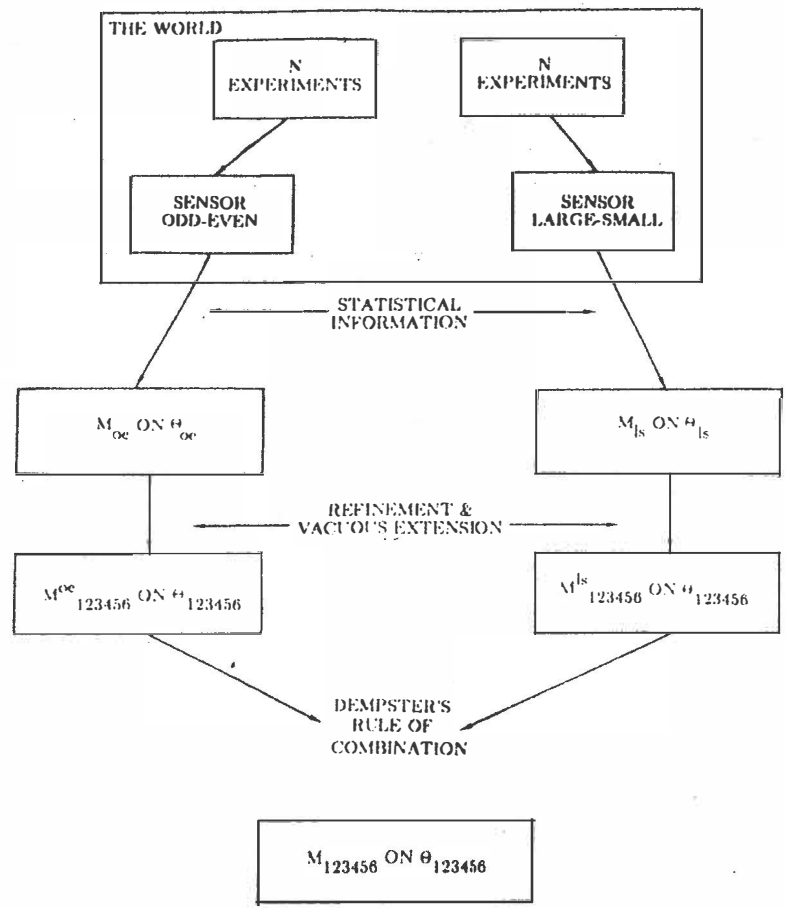

Figure 3: Dempster-Shafer Analysis

odd-even experiment: m(odd) = .5, m(even) = .5

large-small experiment: m(large) = .5, m(small) = .5

The interpretation of the results is that there is no doubt about the die with respect to odd-even and small-large. The probability masses on each set are exactly .5. That is, $Bel(A) = Pl(A) = .5$ for $A \in$ {odd, even, large, small}. This feature places these basic probability assignments in the category of quasi-support functions. These results are intuitive. As the number of trials gets large, one should eventually be content that the die is fair with respect to these attributes.

The second step is the refinement of the two quasi-support functions to a common frame of discernment. This is done by first defining the refining functions $\omega_{oe}$, and $\omega_{ls}$, where:

$\omega_{oe} : 2^{\{odd\ even\}} \to 2^{\{123456\}}$  $\omega_{ls} : 2^{\{large\ small\}} \to 2^{\{123456\}}$

$\omega_{oe} : \{odd\} \to \{135\}$  $\omega_{ls} : \{large\} \to \{456\}$

$\omega_{oe} : even \to \{246\}$  $\omega_{ls} : \{small\} \to \{123\}$

Secondly, we assume that the way to transfer the quasi-support functions to a common frame of discernment is by vacuous extension. The results of this step are:

81

odd-even experiment:   $m(1,3,5) = .5$ , $m(2,4,6) = .5$
large-small experiment:   $m(4,5,6) = .5$ , $m(1,2,3) = .5$

The third step is to combine the refined support functions using Dempster's Rule of Combination. The results are:

$$m(1,3) = m(2) = m(4,6) = m(5) = .25$$

The interpretation of these results is that there is no uncertainty about the probability mass on "2" and "5", i.e., their probability masses are exactly .25 ! This is certainly not intuitive.

If a large but finite number of tosses are considered instead of the limiting case, we can approximate the results of the statistical estimation procedure by:

$$m_{ls}(large) = .5 - \frac{\epsilon}{2} \quad m_{ls}(small) = .5 - \frac{\epsilon}{2} \quad m_{ls}(\Theta) = \epsilon$$

$$m_{oe}(odd) = .5 - \frac{\epsilon}{2} \quad m_{oe}(even) = .5 - \frac{\epsilon}{2} \quad m_{oe}(\Theta) = \epsilon$$

Going through the same process of refinement, vacuous extension, and Dempster's Rule of Combination, we arrive at the results:

$$m(2) = m(5) = m(1,3) = m(4,6) = (.5 - \frac{\epsilon}{2})^2$$

$$m(1,2,3) = m(4,5,6) = m(1,3,5) = m(2,4,6) = (.5 - \frac{\epsilon}{2}) \cdot \epsilon$$

$$m(\Theta) = \epsilon^2$$

The corresponding support functions and plausibility functions for the singleton sets, (ignoring terms of $o(\epsilon^2)$) are:

$$\text{for } i = 1,3,4,6: \quad S(i) = 0 \quad Pl(i) = .25 + \frac{\epsilon}{2}$$

$$\text{for } i = 2,5: \quad S(i) = .25 - \frac{\epsilon}{2} \quad Pl(i) = .25 + \frac{\epsilon}{2}$$

## 5. Analysis of Experimental Results

In this section we analyze the experimental results from metaprobability theory and Dempster-Shafer theory. Although both theories recognize the inherent symmetries of (2, 5) and (1, 3, 4, 6), the similarities between the results are few. In general, metaprobability theory seems better fitted for this type of analysis. This is not surprising and perhaps even a little bit unfair since it is generally recognized that probability theory is well-suited for what is essentially a statistical estimation problem. We will first discuss the metaprobability results and then the Dempster-Shafer results.

We are satisfied with the results of metaprobability and the correctness of our method. The metaprobability results have the interpretation that only first-order distributions which meet the implicit constraints have a positive metaprobability mass. This seems reasonable. The prior distribution does not really play a major role in this analysis since we have considered the limiting case (i.e. amount of evidence goes to infinity). However, the prior meta-distribution does allow us to include our prior beliefs about a "normal" die. If we believe that a "normal" die is "fair", it would be easy to encode this in the prior distribution. Given a prior, the metaprobability analysis will give reasonable results not only for the limiting case but also for any finite amount of evidence. We believe that metaprobability provides solutions to many of the traditional problems with using probability in evidential reasoning including the problem of representing ignorance. A potential drawback is the amount of computation required.

On the other hand, the results of applying Dempster-Shafer theory are counter-intuitive. One of the key attributes of Dempster-Shafer is its ability to represent ignorance. However, application of Dempster's Rule of Combination results in precise beliefs on the events {2 5} when ignorance should still be present. The problem is due to the requirement of Dempster's Rule for distinct and independent bodies of evidence. Although the two bodies of experimental evidence are based on independent observations of independent tosses, the tosses originate from the same underlying source of uncertainty. This violates the "distinct and independent" criteria.

## 5. Conclusions

We believe that some generalization of probabilistic reasoning is necessary for evidential reasoning. Our work has shown that the use of metaprobability is a reasonable generalization. Dempster-Shafer on the other hand, gives counter-intuitive results and needs further clarification with respect to inference.

## Acknowledgements

This work was partially supported by U.S. Army contract DAAK21-84-0015.

## References

[Dempster 1967] Dempster, A.P., "Upper and Lower Probabilities Induced by a Multivalued Mapping," *Annals of Mathematical Statistics*, Vol. 38, (1967).

[Reichenbach 1949] Reichenbach, H., *The Theory of Probability*, (Berkeley, California: University of California Press, 1949).

[Shafer 1976] Shafer, G., "A Mathematical Theory of Evidence," (Princeton University Press, 1976).